%% file: CoG_Paper_new.tex
\documentclass[conference]{IEEEtran}
\IEEEoverridecommandlockouts
\usepackage{cite}
\usepackage{amsmath,amssymb,amsfonts}
\usepackage{algorithmic}
\usepackage{graphicx}
\usepackage{textcomp}
\usepackage{booktabs}
\usepackage[dvipsnames]{xcolor}
\def\BibTeX{{\rm B\kern-.05em{\sc i\kern-.025em b}\kern-.08em
    T\kern-.1667em\lower.7ex\hbox{E}\kern-.125emX}}

\usepackage{xskak}
\usesymfig
\setboardfontsize{11pt}
\usepackage{chessboard}

\usepackage{amssymb}
\usepackage{hyperref}
\usepackage{numprint}
\usepackage{svg}
\usepackage{caption}
\usepackage{graphicx}
\usepackage{subcaption}
\usepackage{todonotes}
\usepackage{comment}
\usepackage{xcolor}

\newlength{\frameRule}
\newlength{\frameSep}
\definecolor{turquoise}{RGB}{64, 224, 208}
\newcommand{\network}{\textsl}

\newcommand{\repair}{\mbox{\textsc{RePAIR}}}

\usepackage{xcolor}

\begin{document}

\title{\repair{}: Predictive Self-Supervised \\ Representation Learning in Chess}

\author{\IEEEauthorblockN{1\textsuperscript{st} Christoph Koller}
\IEEEauthorblockA{\textit{JKU} \\
Linz, Austria \\
ckoller@faw.jku.at}
\and
\IEEEauthorblockN{2\textsuperscript{nd} Johannes Fürnkranz}
\IEEEauthorblockA{\textit{JKU} \\
Linz, Austria \\
juffi@faw.jku.at}
\and
\IEEEauthorblockN{3\textsuperscript{rd} Timo Bertram}
\IEEEauthorblockA{\textit{JKU} \\
Linz, Austria \\
bertram@ml.jku.at}
}

\IEEEoverridecommandlockouts
\IEEEpubid{\makebox[\columnwidth]{979-8-3315-9476-3/26/\$31.00 \copyright2026 IEEE\hfill} %
\hspace{\columnsep}\makebox[\columnwidth]{ }}

\maketitle

\IEEEpubidadjcol

\begin{abstract}
  In this paper, we introduce \textit{Representation Prediction via Autoencoding using Iterative Refinement} (\repair{}) \,---\, a novel self-supervised representation learning architecture that synthesizes Masked Autoencoders (MAE), Joint Embedding Predictive Architectures (JEPA), and Bidirectional Encoder Representations from Transformers (BERT). We demonstrate how it can be used to encode objects in sequential data like consecutive chess positions into compact yet meaningful representations. The basic principle of the architecture is to mask large portions of a sequence of latent states, similar to BERT and MAE. Then, we apply a lightweight \network{Predictor} to the latent representations that repairs gaps in the sequence in a lower-dimensional embedding space akin to JEPA. Our experiments in the domain of chess show that the \network{Encoder} refines the board representations such that meaningful chess concepts emerge clustered in the latent space. Furthermore, reconstructions of the masked board states show that the model is able to reason about the piece movements without relying on costly reinforcement learning methods. Lastly, we find that the resulting representation space allows for quick and intuitive dissections of chess games by observing the game path trajectories in this semantically rich space.
\end{abstract}


\input{main_new}

\newpage

\bibliographystyle{IEEEtran}
\bibliography{references}

\end{document}

%% file: main_new.tex
\section{Introduction}\label{section:Introduction}

Recent developments in %
representation learning, especially in the image domain, have shown that meaningful representations can be learned in self-supervised fashion \cite{I-JEPA, MAE, BYOL, DINO}. However, in the field of chess, most methods rely on either reinforcement learning \cite{AlphaZero, lc0}, or domain expertise and handcrafted heuristics \cite{Stockfish}. As an alternative, this paper proposes a method for learning %
representations of states from sequential data in a self-supervised fashion. The architecture is tested by mapping consecutive chess states into a rich low-dimensional space. In particular, we take inspiration from the Joint Embedding Predictive Architecture (JEPA) \cite{I-JEPA, lecun2022path}, Masked Autoencoders (MAE) \cite{MAE} as well as the workflow of Bidirectional Encoder Representations from Transformers (BERT) \cite{BERT}.

The key idea is to take a sequence of objects or states, e.g., a %
game of chess, and mask parts of the sequence. Then, an \network{Encoder} maps the incomplete sequence into a low-dimensional latent space, followed by a lightweight \network{Predictor} that is trained to repair the latent sequence iteratively. Lastly, the latent representations are mapped back to the original space, where the full sequence is recovered. By training the model to reconstruct missing %
states in the sequence, the \network{Encoder} is forced to learn the structure and semantics of the data. This enables the repair of large gaps in the latent sequence, resulting in a semantically rich representation space.
We term the resulting workflow \textit{Representation Prediction via Autoencoding using Iterative Refinement} (\repair{}).

To demonstrate the quality of the reconstructions, we show the top-1 reconstruction accuracy of the architecture and provide multiple examples that illustrate how the model can infer probability distributions for reconstructed states across various gap sizes. Furthermore, we show the versatility and generality of the resulting representations using several different chess datasets. First, in a Lichess dataset of full games, we identify separate regions in the embedding space that capture different phases of the games, such as opening, castling, middle game, and endgame.
For the set of ECO opening positions, we observe a spatial clustering according to common opening choices. In the Lichess chess puzzle dataset, we find that positions with common positional or combinatorial motifs are clustered %

To summarize, our paper demonstrates the following:
\begin{itemize}
    \item Our \network{Encoder} can map chess positions into a semantically meaningful %
    space, where chess concepts form clusters.
    \item Our \network{Predictor} can infer chess moves in this latent space.
    \item The model can reason about the basics of chess without resorting to handcrafted heuristics or expensive reinforcement learning methods.
    \item Large gaps in a sequence of chess states can be partially or fully restored using \repair{}.
    \item Chess games can be intuitively dissected by analyzing their path in the learned latent space.
\end{itemize}

In the following, we start with a discussion of related work in representation learning with a particular focus on chess (\autoref{section:Related work}). Then, we give a general description of the proposed \repair\ framework (\autoref{section:Method}) and its realization in a chess context (\autoref{section:Chess setup}). Lastly, the results of the experiments, including a detailed analysis of the learned representations on several chess datasets can be found in \autoref{section:Results}, before we conclude.

All our code and the learned embeddings can be accessed at 
\url{https://github.com/Artificial-Chrisi/RePAIR}.

\section{Related work}\label{section:Related work}

The following section is split into two parts, differentiating between general representation learning architectures and chess-specific methods.

\subsection{Representation Learning Architectures}\label{subsection:Related Architectures}
This paper takes inspiration from the self-supervised learning methods BERT \cite{BERT}, MAE\cite{MAE}, and Image-JEPA\cite{I-JEPA}. BERT partially masks sequences of tokens, i.e., sentences, and predicts a distribution over the dictionary for masked positions. By applying the model autoregressively and sampling from its output distribution, the model can generate new sentences given any starting sequence. MAEs mask regions of images and map the visible regions into a low-dimensional space. Afterwards, a decoder collects all latent patches of a given image with their locations and aims to reconstruct the entire image. Lastly, Image-JEPA processes images similar to MAE. An image is cropped and partially masked, with its masked regions mapped into a learned space. Instead of using a decoder, the unmasked image is projected into the same space, where the two representations are aligned by predicting the unmasked representation from the masked representation. %

As our task is reconstruction rather than latent alignment, the architecture is more similar to BERT and MAE than to JEPA. As with BERT, our architecture predicts from a masked sequence, but we directly predict latent states given by the \network{Encoder} instead of token distributions. This is similar to MAEs with the key difference that only the \network{Predictor} receives temporal information. In MAEs, the latent representations are all collected and processed by the \network{Decoder}, whereas in \repair{}, each latent state is decoded independently.%

\subsection{Learning Chess Representations}\label{subsection:Chess}

Our primary goal is learning a semantically rich representation space for chess positions. Previous work has aimed to achieve a similar goal with supervised contrastive learning \cite{2025_constrastive_chess}, but requires labeled positive and negative samples. %
Chess2Vec \cite{Chess2Vec} learns the relation between positions and commonly played moves.
Chess representations can also be extracted from trained chess agents such as AlphaZero \cite{AlphaZero}, Leela Chess Zero \cite{lc0} or Stockfish \cite{Stockfish}. Training such agents requires costly reinforcement learning and large-scale self-play setups. Agents trained with supervised learning, such as Maia \cite{Maia-2}, can be used in a similar fashion.

In contrast to the above, \repair{} strictly relies on self-supervised latent prediction without incorporating explicit move information, player Elo, engine-generated labels, or reinforcement learning.

\section{Method}\label{section:Method}
The following section describes \repair{}, which receives a sequence of states and learns a representation space in self-supervised manner. For simplicity, we henceforth assume a sequence of states, but in general, the architecture can be applied to any sequence of objects that undergo a successive transformation.

\begin{figure}[h!]
    \centering
    \includegraphics[width=1.0\linewidth]{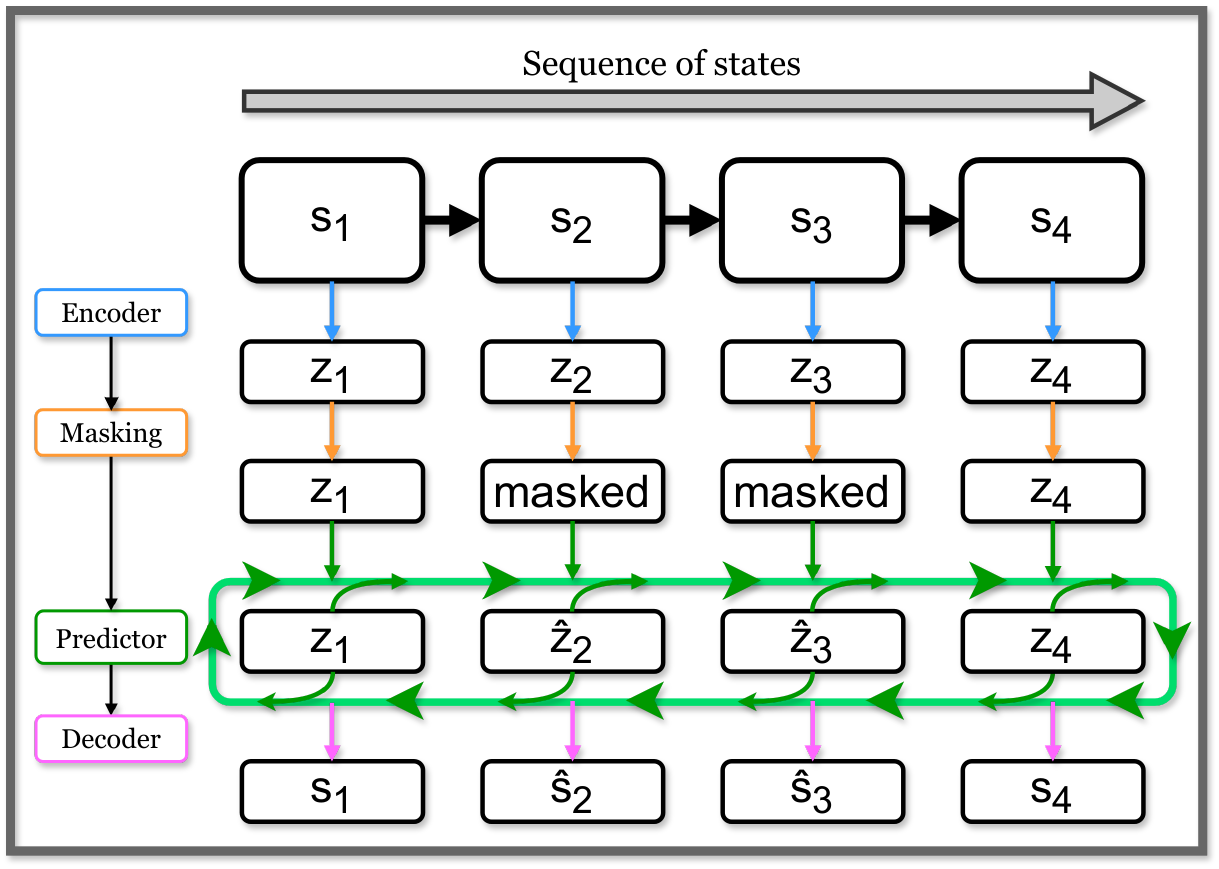}
    \caption{High-level overview of \repair{}. States $s_1$--$ s_4 \in \mathcal{S}$ are first mapped independently into latent states $z_1$--$z_4 \in \mathcal{Z}$ by the \network{Encoder} (\textcolor{SkyBlue}{blue}). $z_2$ and $z_3$ are masked (\textcolor{orange}{orange}). The \network{Predictor} (\textcolor{ForestGreen}{green}) repairs the sequence via iterative refinement, using an attention mechanism. Then, the states are decoded independently (\textcolor{magenta}{pink}).
    }
    
    \label{fig:architecture}
\end{figure}

\autoref{fig:architecture} depicts the architecture, which consists of an \network{Encoder}, a \network{Predictor}, and a \network{Decoder}, with an intermediate masking step.
The states $s_1$--$s_4$ 
are mapped to latent states $z_1$--$z_4$ by the \network{Encoder}  (\autoref{subsection:Encoder}). Next, randomly selected parts of the sequence ($z_2$ and $z_3$) are masked  (\autoref{subsection:Masking}). The \network{Predictor} iteratively refines the latent representations by processing the entire sequence and restoring the masked states  (\autoref{subsection:Predictor}). Lastly, the restored states are independently mapped to the input space using the \network{Decoder}  (\autoref{subsection:Decoder}).

\vspace{\baselineskip}

\subsection{Encoder}\label{subsection:Encoder}

The \network{Encoder} transforms initial states into latent representations that can be repaired by the \network{Predictor} (\autoref{subsection:Predictor}). Formally, each state $s_i$ in a sequence of states $\mathcal{S} = \{s_1, \dots, s_n\}$ is mapped to a compact vector $z_i =$ \network{Encoder}$(s_i)$, resulting in a sequence of vectors $\mathcal{Z}$.

Our chess-specific architecture starts with a single convolution that transforms the input to have the desire number of latent channels.
A series of Squeeze-and-Excitation blocks \cite{SqueezeAndExcitation} is applied, which use a lightweight gating mechanism to learn a weight for each channel. This allows the network to emphasize important features. Lastly, the latent channels are reduced by a final convolution and the result is flattened to a compact vector. The \network{Encoder} is applied to each state independently with no information flow in the sequence. %

\subsection{Masking}\label{subsection:Masking}

After transforming a state sequence $\mathcal{S}$ into a feature vector sequence $\mathcal{Z}$, random parts of this sequence are masked. For this, with probability $p$, every latent vector $z_i \in \mathcal{Z}$ except for the first and last element of $\mathcal{Z}$
is set to a learnable masking token $\bar{z}_i = \network{Masking}(z_i,p)$%
, resulting in a partially masked sequence $\mathcal{\bar{Z}}$. The choice of $p$ depends on the task: natural language domains require small gaps with larger amounts of context, while chess games allow for larger gaps.

\subsection{Predictor}\label{subsection:Predictor}

The \network{Predictor} is the only part of the architecture that allows for information exchange between states in the sequence. The predictor is applied iteratively over the entire sequence and updates masked states using an attention mechanism.

More precisely, the \network{Predictor} processes all latent states $\bar{z}_i \in \mathcal{\bar{Z}}$ to generate an updated representation $\hat{z}_i = $ \network{Predictor}($\bar{z}_{i}, \mathcal{\bar{Z}}$) %
resulting in the sequence $\hat{\mathcal{Z}}^{(1)}$. %
The \network{Predictor} is applied repeatedly, resulting in iteratively refined states $\hat{z}_i^{(j)} =$ \network{Predictor}($\hat{z}_i^{(j-1)}, \hat{\mathcal{Z}}^{(j-1)})$, where $j$ represents the $j$-th iteration of this prediction step. After $n$ prediction steps, the goal is to reach $z_i \approx \hat{z}_i^{(n)} = $ \network{Predictor}$(\hat{z}_i^{(n-1)}, \hat{\mathcal{Z}}^{(n-1)})$, i.e., the latent representation of each masked state should converge to the unmasked ground truth.

Consequently, we need sufficient layers such that the model is capable of restoring long sequence. However, it is desired to keep the \network{Predictor} small, so that the \network{Encoder} performs the majority of the work. For the given task, all layers perform the same task: processing a sequence of states and predicting an improved version of the sequence. Therefore, we use a single-layer Transformer and apply it repeatedly on the state sequence \cite{Takase2021, Dabre2019}. The Transformer follows an encoder-only architecture with four heads.

\subsection{Decoder}\label{subsection:Decoder}

After the \network{Predictor} has repaired the masked sequence, we decode the latent representations back to the input space. The \network{Decoder} is simply an inverted \network{Encoder}, i.e., we reshape the vector from 1D to 3D and use convolutional layers and Squeeze-and-Excitation blocks. The goal is to achieve $s_i \approx\hat{s}_i = $ \network{Decoder}$(\hat{z}_i)$, i.e., to reconstruct the original elements of the sequence. As a consequence, we obtain a learned mapping from the semantically rich representation space into the input space, where latent states can be interpreted as chess positions again.

\subsection{Losses}\label{subsection:Loss}

As previously outlined, the task is to repair a sequence of states by encoding and decoding in a state-by-state manner, including a lightweight \network{Predictor} that performs the restoration in latent space. This process yields multiple objectives we can optimize.

\begin{figure}[htbp]
    \centering
    \includegraphics[width=1.0\linewidth]{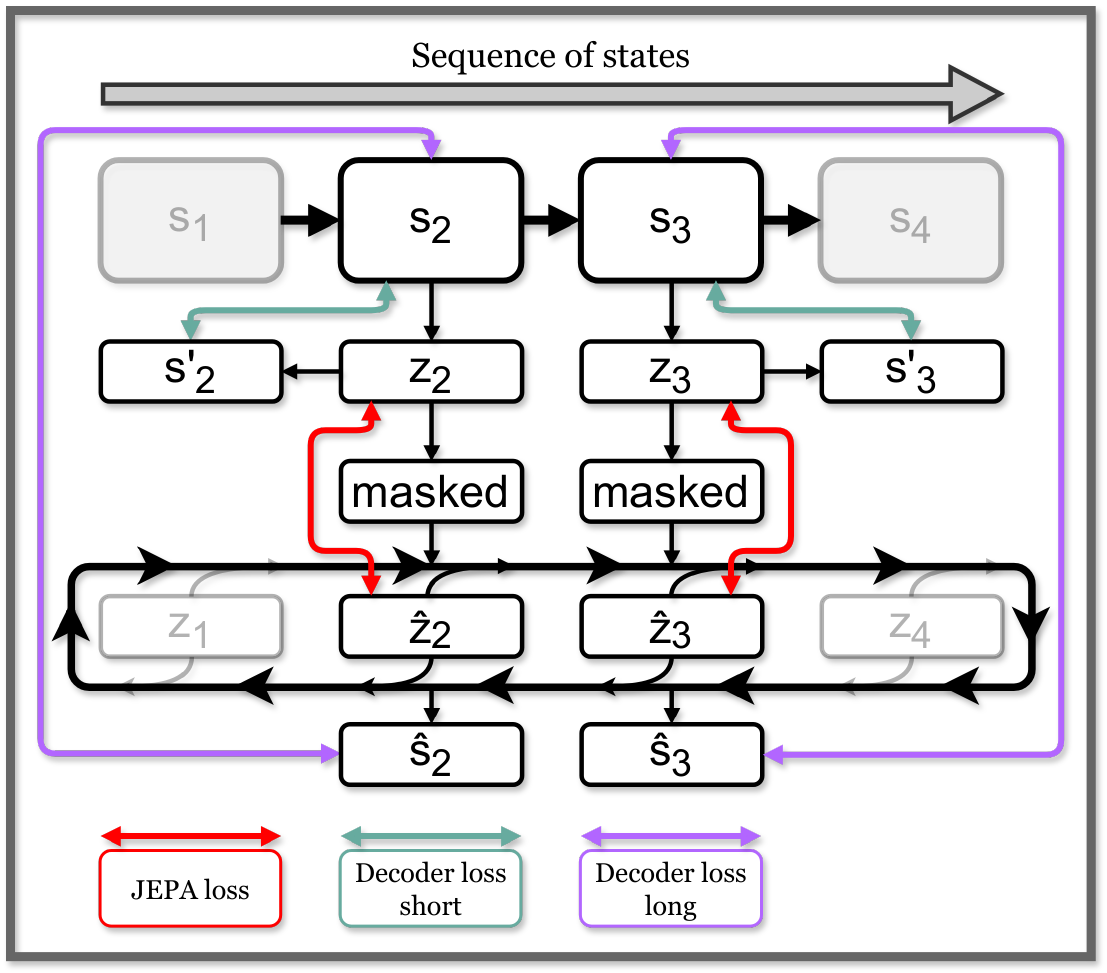}
    \caption{The loss functions of \repair{}. The full architecture is described in \autoref{fig:architecture}. The loss between latent representations (\textcolor{red}{red}) and the loss of the state reconstructions can be optimized. The \network{Decoder} can be applied both on the input as well as the output of the \network{Predictor}, yielding a short (\textcolor{teal}{teal}) and a long (\textcolor{violet}{violet}) \network{Decoder} path.}
    
    \label{fig:architecture_loss}
\end{figure}

\autoref{fig:architecture_loss} depicts the architecture of \autoref{fig:architecture} with possible objectives included. The initial state $s_1$ as well as $s_4$ are omitted as the loss is only computed on masked states. In the following, we discuss the three losses in detail. An evaluation of these choices can be found in \autoref{section:Loss performance}.

\paragraph{JEPA loss}
The first objective we could optimize is the loss between the \network{Predictor} output and the unmasked latent representation, e.g., by computing the MSE between the feature vectors:
$$
\mathcal{L_\textbf{JEPA}} = (z_i - \hat{z}_i)^2
$$
where $z_i$ and $\hat{z}_i$ are the latent board representations at sequence location $i$ before and after the \network{Predictor} step. This corresponds to the latent alignment objective in JEPA architectures and will therefore be referred to as the JEPA loss. 

\paragraph{Decoder Loss\,---\,long}
JEPA architectures can suffer from representation collapse where all inputs collapse into a single encoding. This can be circumvented by adding explicit regularization, which enforces high variance along all encodings of a batch \cite{VCReg}, or with a bootstrap-style \network{Encoder} \cite{I-JEPA, BYOL}.
In our setting, the \network{Decoder} enforces diverse representations, as it would otherwise yield collapsed decodings. Integrating the \network{Decoder} into the optimized objective, e.g., with a cross entropy loss between the states in the input sequences and the reconstructions thus prevents representation collapse.

Consequently, the decoder loss along the long path of the network is denoted as 

$$
\mathcal{L_\textbf{DL}}(s_i, \hat{s}_i)
$$
where $s_i$ is the state at sequence position $i$ and $\hat{s}_i$ is the decoded \network{Predictor} output at position $i$.
Concrete instantiations of this loss depend on the domain\,---\,our choices in the domain of chess will be discussed in \autoref{section:Loss function}.

\paragraph{Decoder Loss\,---\,short}
Alternatively, we can encode and immediately decode each state, which we call the short \network{Decoder} path:

$$
\mathcal{L_\textbf{DS}}(s_i,s'_i)
$$
where $s_i$ is the state at sequence position $i$ and $s'_i$ is the \network{Decoder} output at position $i$ without applying the masking procedure and the \network{Predictor}, i.e., $s'_i = $ \network{Decoder}($z_{i})$. This loss does not optimize the \network{Predictor}. Therefore, we need to combine it with the JEPA loss to ensure optimization of the entire model.

\autoref{fig:architecture_loss} shows the distinction between the short (\textcolor{teal}{teal}) and the long (\textcolor{violet}{violet}) \network{Decoder} path. Our choices for chess sequence reconstructions are discussed in \autoref{section:Loss function}.

\section{Chess setup}\label{section:Chess setup}

The following section outlines how we apply \repair{} to chess. 

\paragraph{Chessboard representation}\label{section:Input representation}
Each chess state is represented as a $19\times 8\times 8$ bitmap. The first 12 channels are the locations of the 6 unique chess pieces for the respective color, followed by channel 13, which encodes empty squares. Channel 14 to 17 specify the respective castling rights, and the last two channels indicate possible en passant squares and the current player.
Previous models frequently use a $18\times 8\times 8$ board representation, as they omit the "empty square" channel \cite{Maia-2}. We use this additional channel such that we can view reconstructed boards as probability distributions over all possible options for each square. 
Complete chess games are represented as a sequence of chess states, where each sequence is truncated or padded to 100 elements. This leads to an overall representation of chess games as $100 \times 19 \times 8 \times 8$ bitmap tensors.

\paragraph{Positional Encoding of Board Sequences}\label{section:Positional Encoding}

When encoding the positions of states in the sequence, the relative distance between boards is more important than the absolute position in the game.
We thus apply RoPE \cite{RoPE}, as it allows the model to be agnostic towards the global position of a state and to only consider relative distances between encodings. This also allows us to extend games beyond 100 moves at inference, as the model can generalize beyond positions that were encountered during training.

\paragraph{Loss function}\label{section:Loss function}
We use the categorical cross entropy loss along the first 13 channels of the bitmaps as the optimized objective for the \network{Decoder} losses  $\mathcal{L_\textbf{DS}}(s_i,s'_i)$ and $\mathcal{L_\textbf{DL}}(s_i, \hat{s}_i)$. For the remaining channels, we use the binary cross entropy loss between each square. Optimizing both loss functions jointly yields the reconstruction of the entire $19\times 8\times 8$ bitmap. We provide an ablation over the combination of objectives in \autoref{section:Loss performance}.

\paragraph{Data}\label{section:Data}

The model was trained on \numprint{80000} games of the Lichess database (June 2025).\footnote{\url{https://database.lichess.org/}} 
There was no restriction on game length or Elo rating, as the model should learn chess semantics across all players. Additionally, \numprint{10000} games are each used as validation and test sets.

\paragraph{Training}\label{section:Training}

The model was trained until convergence with a masking ratio of 50\%, followed by a second training phase using 90\% masking, resulting in expected average gap sizes of approximately 2 and 10 states, respectively. BERT uses a masking ratio of 15\%, but in language, a large context is needed to fill in small gaps, whereas in chess, small amounts of context allow for large information inference. A small masking ratio would trivialize the task, whereas a too-large masking ratio can be too difficult to learn.

\section{Results}\label{section:Results}
We begin by comparing the combination of objectives to optimize in training \mbox{(\autoref{section:Loss performance})}, followed by a discussion of the reconstruction performance \mbox{(\autoref{section:State reconstruction})}, and an analysis of the network's attention behavior \mbox{(\autoref{section:Attention})}. 
Finally, we analyze the semantics of the learned representation space using multiple datasets that capture various aspects of the game \mbox{(\autoref{section:Representation Space})}.

\subsection{Loss performance}\label{section:Loss performance}

We compare the performance of different \repair{} models trained with the losses discussed in \autoref{subsection:Loss}.
For this, we only use the five meaningful combinations; using only $\mathcal{L_\textbf{DS}}$ or only $\mathcal{L_\textbf{JEPA}}$ would not train the \network{Predictor} or the \network{Decoder} and is thus discarded.

Each of the five models was trained using a static masking ratio of 80\% for three separate runs. We do not split the training into two phases for ease of computation (see \autoref{section:Method}). For each model, we compute the average \mbox{top-1} accuracy for each possible element, i.e., all pieces and an empty square, over all squares. 
\autoref{table:top1acc} shows the results for the different loss functions.

\begin{table}[h]
\centering
\begin{tabular}{cc}
\toprule
Loss & Top-1 Accuracy\\
\midrule
Lower bound (predict empty) & 50.00\% \\
\midrule
$\mathcal{L_\textbf{DS}} + \mathcal{L_\textbf{JEPA}}$ & 73.41\% $\pm$ 0.42\% \\
$\mathcal{L_\textbf{DS}} + \mathcal{L_\textbf{DL}}$ & 93.71\% $\pm$ 0.05\% \\
$\mathcal{L_\textbf{DL}}$ & 93.90\% $\pm$ 0.02\% \\
$\mathcal{L_\textbf{DL}} + \mathcal{L_\textbf{JEPA}}$ & 93.18\% $\pm$ 0.02\% \\
$\mathcal{L_\textbf{DS}} + \mathcal{L_\textbf{DL}} + \mathcal{L_\textbf{JEPA}}$ & 93.19\% $\pm$ 0.02\% \\
\bottomrule
\end{tabular}
\caption{Top-1 accuracy for various combinations of loss functions. The lower bound corresponds to always predicting empty squares (50\%). Not including the long decoder loss $\mathcal{L_\textbf{DL}}$ results in significantly lower accuracy, while all other versions perform similarly.}
\label{table:top1acc}
\end{table}

All models that include $ \mathcal{L_\textbf{DL}} $ in their loss function perform similarly, correctly reconstructing 93\% of all squares despite masking 80\% of all states in the sequence. Only the combination of $\mathcal{L_\textbf{DS}}$ and $\mathcal{L_\textbf{JEPA}}$ performs poorly because the \network{Predictor} collapses. %
However, since every chess game has at most 32 pieces on the 64 square board, simply predicting every square to be empty already yields a top-1 accuracy lower bound of 50\%.

\begin{figure}[h]
\centering

\centering\includegraphics[width=0.9\linewidth]{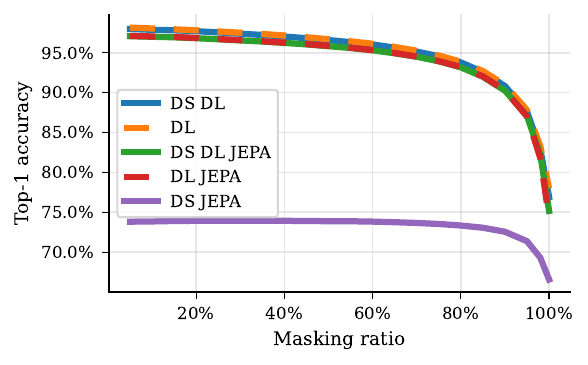}

\caption{Top-1 accuracy of different loss configurations across varying masking ratios in testing. All models are trained with 80\% masking. All combinations except $\mathcal{L_\textbf{DS}} +  \mathcal{L_\textbf{JEPA}}$ perform similarly.}

\label{fig:top1acc_mean}
\end{figure}

To get a more fine-grained comparison of the performance of the models, we assess the accuracy in relation to different masking ratios. We vary the ratio of masked states in the input sequence and compare the resulting top-1 accuracies (\autoref{fig:top1acc_mean}).

Again, the four models that include $ \mathcal{L_\textbf{DL}} $ perform similarly and behave nearly identical across all masking ratios. A significant drop can only be observed once the masking ratio approaches 100\%, at which point the model only sees the start and the end of the sequence.
Since all $ \mathcal{L_\textbf{DL}} $ loss combinations perform similarly, we choose $ \mathcal{L} = \mathcal{L_\textbf{DS}} + \mathcal{L_\textbf{DL}} $ for the remaining experiments, omitting $\mathcal{L_\textbf{JEPA}}$. This allows us to decode states before and after the \network{Predictor} step and imposes fewer restrictions to the \network{Predictor} itself.

\subsection{State reconstruction}\label{section:State reconstruction}
As the \network{Decoder} output can be interpreted as a probability distribution for each square, the resulting output can be visualized as a chess board with superimposed pieces, where the boldness of each piece is proportional to the given probability.

\begin{figure}[h]
    \centering

    \begin{subfigure}{0.3\linewidth}
        \centering
        \includegraphics[width=\linewidth]{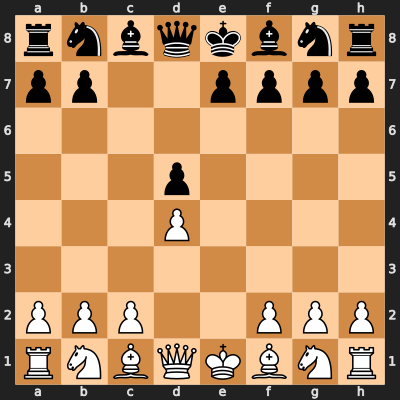}
    \end{subfigure}
    \hfill
    \begin{subfigure}{0.325\linewidth}
        \centering
        \makebox[\linewidth][c]{%
            \fcolorbox{turquoise}{yellow}{%
                \includegraphics[
                    width=\dimexpr\linewidth-2\frameRule-2\frameSep\relax
                ]{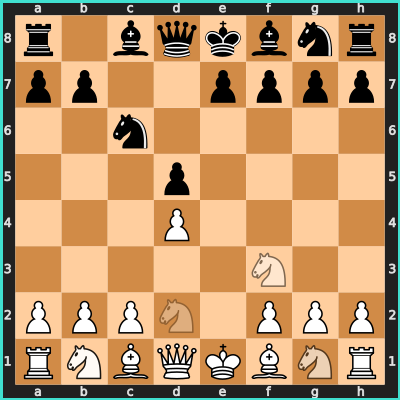}%
            }%
        }
    \end{subfigure}
    \hfill
    \begin{subfigure}{0.3\linewidth}
        \centering
        \includegraphics[width=\linewidth]{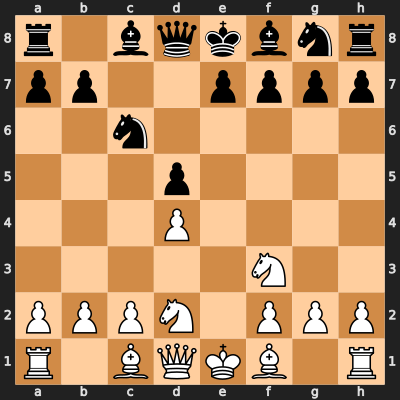}
    \end{subfigure}

    \vspace{0.5em}
    \hrule
    \vspace{0.5em}

    \begin{subfigure}{0.3\linewidth}
        \centering
        \includegraphics[width=\linewidth]{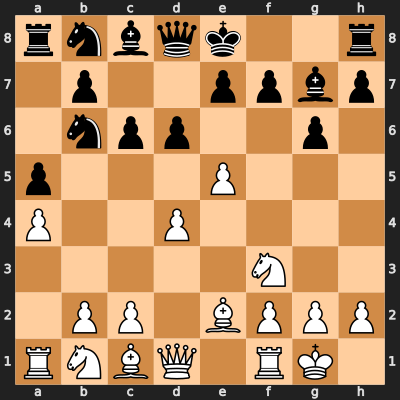}
    \end{subfigure}
    \hfill
    \begin{subfigure}{0.325\linewidth}
        \centering
        \makebox[\linewidth][c]{%
            \fcolorbox{turquoise}{yellow}{%
                \includegraphics[
                    width=\dimexpr\linewidth-2\frameRule-2\frameSep\relax
                ]{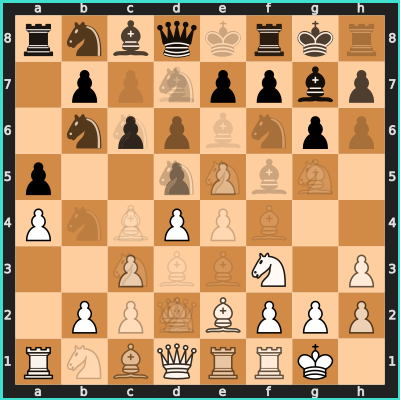}%
            }%
        }
    \end{subfigure}
    \hfill
    \begin{subfigure}{0.3\linewidth}
        \centering
        \includegraphics[width=\linewidth]{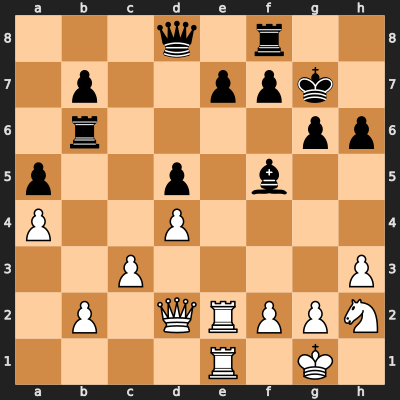}
    \end{subfigure}

    \vspace{0.5em}
    \hrule
    \vspace{0.5em}

    \begin{subfigure}{0.3\linewidth}
        \centering
        \includegraphics[width=\linewidth]{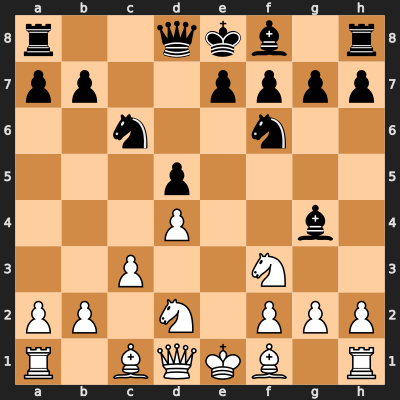}
    \end{subfigure}
    \hfill
    \begin{subfigure}{0.325\linewidth}
        \centering
        \makebox[\linewidth][c]{%
            \fcolorbox{turquoise}{yellow}{%
                \includegraphics[
                    width=\dimexpr\linewidth-2\frameRule-2\frameSep\relax
                ]{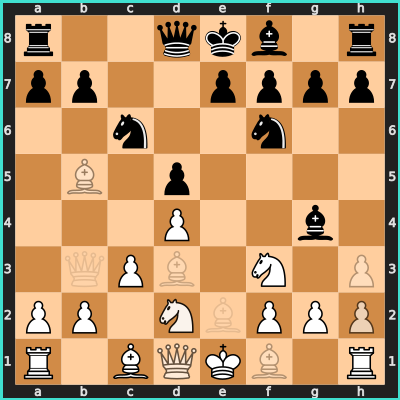}%
            }%
        }
    \end{subfigure}
    \hfill
    \begin{subfigure}{0.3\linewidth}
        \centering
        \includegraphics[width=\linewidth]{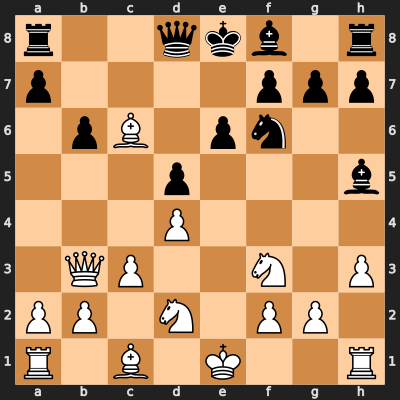}
    \end{subfigure}

    \vspace{0.5em}

    \caption{Reconstructions of masked game states in three sequences of chess positions. In each row, the left and right boards are given, while the center shows a missing board that is repaired and decoded.  
    The three rows illustrate a simple interpolation (top), the reconstruction of legal move patterns in long gaps (center), and the exclusion of non-sensible moves (bottom). 
}
    \label{fig:reconstruction}

\end{figure}

\begin{figure*}[t!]
\centering
\begin{subfigure}[c]{0.67\linewidth}
    \centering
    \includegraphics[width=0.49\linewidth]{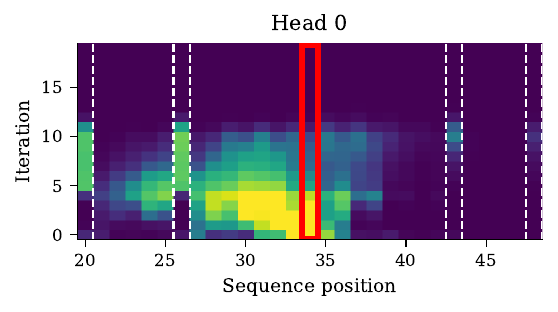}
    \includegraphics[width=0.49\linewidth]{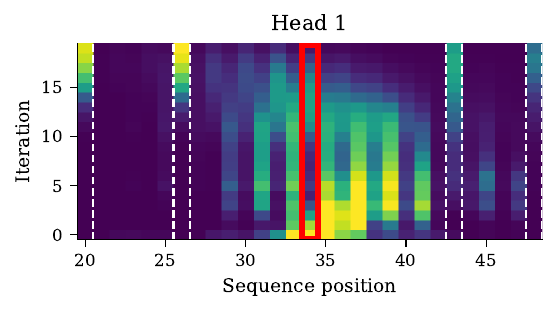}
    \includegraphics[width=0.49\linewidth]{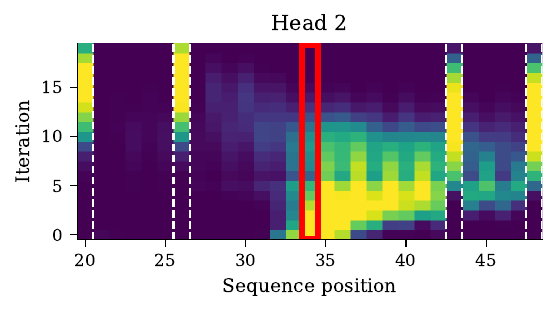}
    \includegraphics[width=0.49\linewidth]{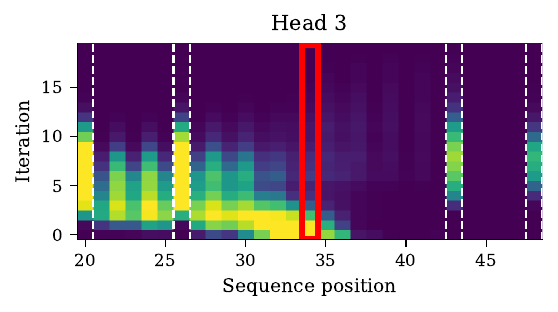}
    \caption{The attention weights of the \network{Predictor's} network heads across different iteration steps. The red position indicates the location of the query state. The white lines mark the unmasked anchor locations. Yellow and purple indicate high and low attention weights respectively. The colors are displayed on a logarithmic scale.}
    \label{fig:attention_heads}
\end{subfigure}
\hfill
\raisebox{0.79cm}{
\begin{subfigure}[t]{0.30\linewidth}
    \centering
    \includegraphics[width=\linewidth]{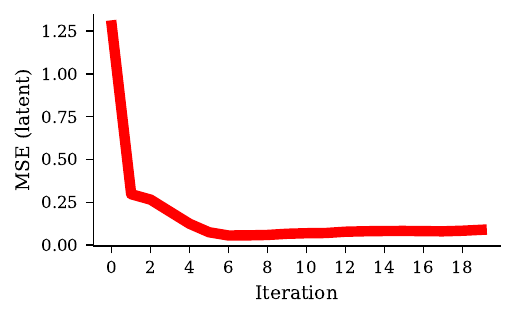}

    \includegraphics[width=\linewidth]{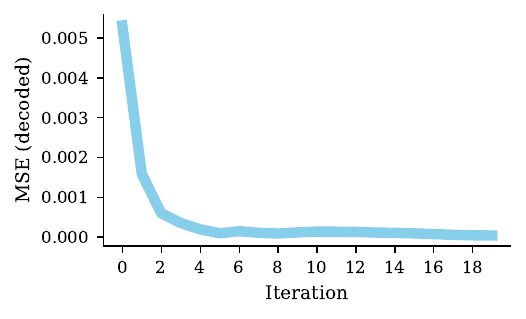}

    \caption{The mean squared difference between consecutive \textcolor{red}{latent} states and consecutive \textcolor{cyan}{decoded} states after every repair iteration.}
    \label{fig:predictor_change}
\end{subfigure}
}
\caption{The results of the \network{Predictor}, visualizing the attention (\autoref{fig:attention_heads}) and the MSE between successive repair iterations over time (\autoref{fig:predictor_change})}
\label{fig:predictor_attention}
\end{figure*}

\autoref{fig:reconstruction} shows the predictions of the model in three selected sample games. Each row depicts a state sequence, where the left board state and the right board state are unmasked and separated by several moves. The center board is one masked board in the sequence, which is repaired and reconstructed. Note that the model obviously predicts as many intermediate board states as there are gaps in the sequence, but for better visualization, we only show one of these states.

The first row depicts a simple example of two possible underlying sequences. To transition from the left board state to the right board state, the sequence must follow \symknight f3, \raisebox{-1.5pt}{\BlackKnightOnWhite}c6, \symknight bd2, or 
\symknight d2, \raisebox{-1.5pt}{\BlackKnightOnWhite}c6, \symknight gf3.
Consequently, the model produces a pseudo-state that shows each of the two knights on
their target square with equal probability. 
The subsequent black knight move is not ambiguous and is therefore correctly detected.

The second row depicts a complex reconstruction with a 25-move gap between the left and right positions. Here, the model considers all possible scenarios. It correctly recognizes the fixed pawns, and the piece probabilities on the other squares all correspond to 
possible move patterns. %
Therefore, the model demonstrates an intuitive understanding of the rules of chess.

The third row depicts an example in which the model demonstrates the ability to reason. In the given sequence, we see three actions. White moves \symqueen b3, attacks \raisebox{-1.5pt}{\BlackBishopOnWhite}g4 with \sympawn h3, and pins \raisebox{-1.5pt}{\BlackKnightOnWhite}c6 with \symbishop b5. After one simulated move, it can be seen that the model takes all three actions into consideration. Moreover, it also considers the possible bishop moves \symbishop e3 and \symbishop d3, while it ignores the moves \symbishop c4 and \symbishop a6.
The latter two moves would lead to an immediate capture of the bishop by a black pawn, and since the bishop delivers the game-ending fork many moves later, the model does not consider these squares.

These examples show that the model not only reconstructs sequences but also takes the ambiguity and the rules of the game into consideration to draw logical conclusions about piece movements.

\subsection{Visualizing the Attention}\label{section:Attention}

As mentioned in \autoref{subsection:Predictor}, the only part of the architecture that communicates between states in the sequence is the \network{Predictor}, as the \network{Encoder} and the \network{Decoder} process each state independently. We thus analyze the attention weights of the \network{Predictor} to gain insights into the communication process between states.

In the following experiment, the one-layer \network{Predictor} is applied 20 consecutive times. \autoref{fig:attention_heads} depicts the attention weights of the four network heads across all iterations, where the query state is marked in red. The unmasked states in the sequence are shown with white lines. %
The color intensity indicates the log-scaled attention of each position to the query vector, where yellow and purple correspond to high and low attention, respectively. Note that we only show elements 20-48 in the sequence, while the model attends to all states.

In the first iteration, all attention is placed on neighboring states. Afterwards, the attention spreads from the masked neighboring states to the unmasked states. The heads focus on different aspects of the sequence:  Head 1 focuses on the communication between neighboring states, while Head 2 and Head 3 focus on propagating information to the left and right states.

\autoref{fig:predictor_change} shows the average squared difference between states while repairing. The change in the states stagnates after roughly five iterations. Comparing this to \autoref{fig:attention_heads}, we see that the fifth iteration corresponds to the moment where communication between neighbors diminishes.

It should also be noted that the attention heatmap shows an alternating horizontal pattern, which could be due to the fact that chess is a two player turn based game, so the heads may attend to one player in particular. Interestingly, we see that attention is put on states beyond the nearest unmasked board despite containing redundant information. To prevent this, we trained another model with no attention outside of the relevant window but no notable improvements were achieved.

\subsection{Representation Space}\label{section:Representation Space}

We visualize the representation space of the \network{Encoder} by projecting it to 2 dimensions using PCA \cite{PCA} and t-SNE \cite{t-SNE}. We show our results on three different datasets:

\begin{enumerate}
    \item A Lichess dataset of full games (December 2025)\footnote{\url{https://database.lichess.org/}}
        \item The ECO chess openings positions\footnote{https://huggingface.co/datasets/Lichess/chess-openings}
    \item The Kaggle chess puzzle dataset\footnote{https://www.kaggle.com/datasets/tianmin/lichess-chess-puzzle-dataset}
\end{enumerate}

This allows us to inspect game trajectories, opening encodings, and puzzle-specific clusters and thus shows the architecture's utility on a wide range of tasks.

\subsubsection{Lichess dataset}\label{section:Lichess dataset}

We start by analyzing entire game trajectories in the representation space, as the path of each sequence yields insights into the semantic structure of our learned space. For visualizing entire games, we use the Projection Space Explorer \cite{PSE}, which allows for dynamic navigation through predefined embedding spaces.

\begin{figure}[h]
\begin{subfigure}{0.48\linewidth}
    \centering
    \includegraphics[width=\linewidth]{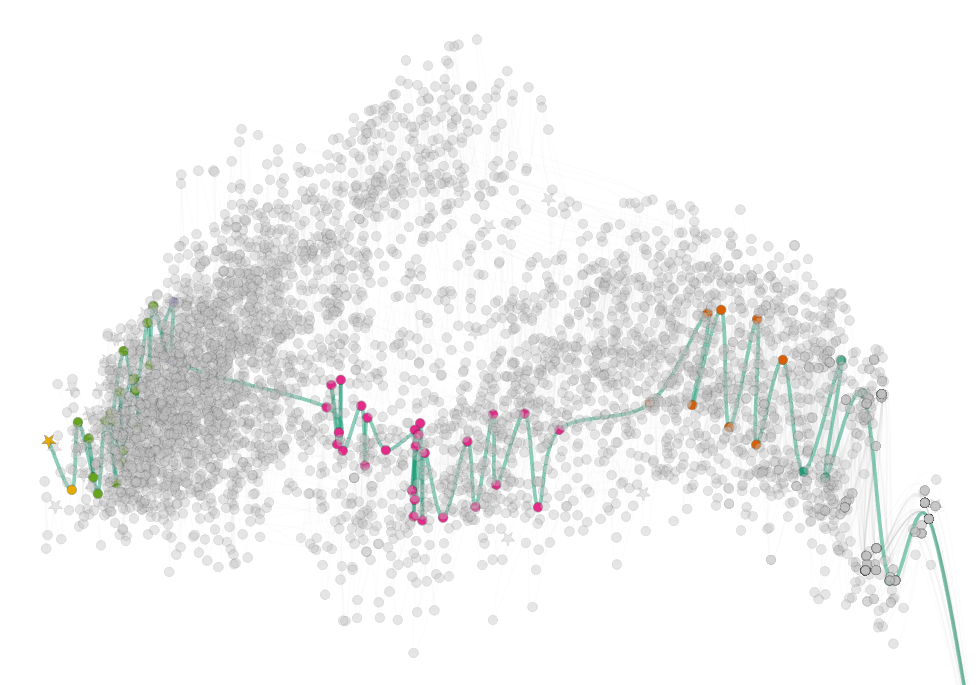}
    \caption{A single trajectory from opening (right) to 
    endgame (left).
    The locations of the states give an immediate impression of the game.}
    
    \label{fig:pse_line}
\end{subfigure}
\hfill
\begin{subfigure}{0.48\linewidth}
    \centering
    \includegraphics[width=0.93 \linewidth]{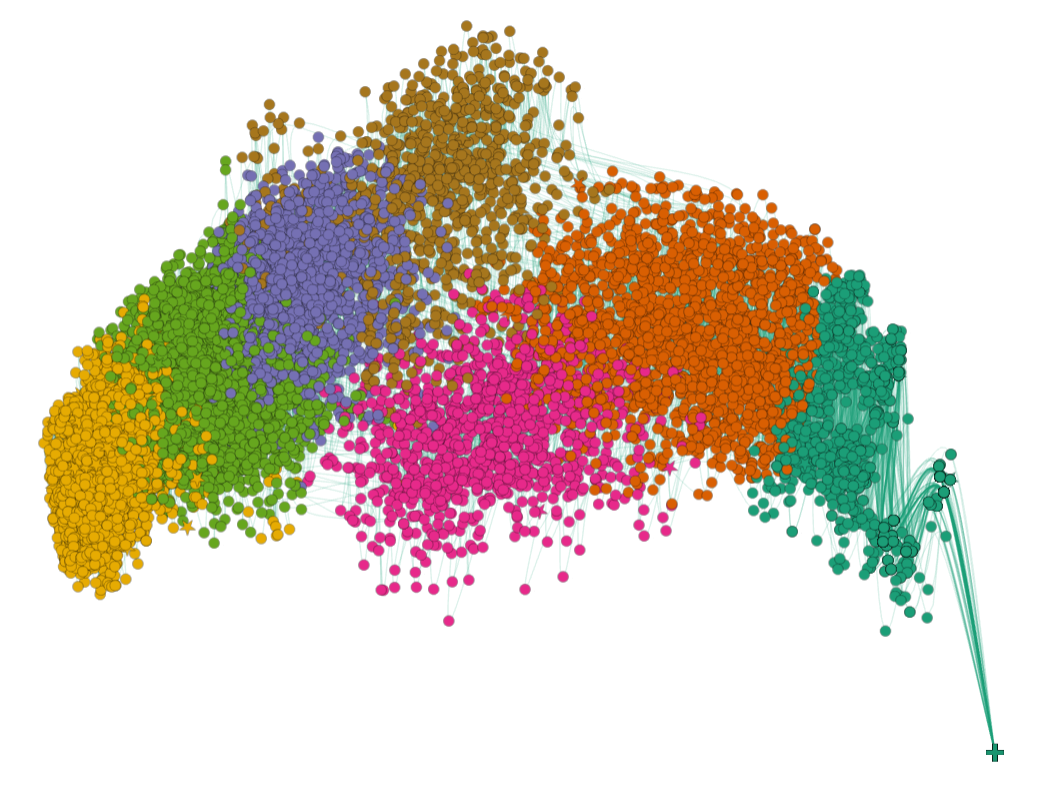}
    \caption{Embedding space colored with seven clusters given by k-means. These clusters correspond to different game phases.}
    \label{fig:pse_cluster}
\end{subfigure}
\caption{Games from the Lichess dataset projected into the representation space and reduced to 2 dimensions using PCA.}
\label{fig:pse}
\end{figure}

\autoref{fig:pse} shows the learned embeddings projected to 2 dimensions using PCA. In total, \numprint{10000} games were projected and a subset of \numprint{200} games was selected and visualized using the Projection Space Explorer\footnote{\url{https://jku-vds-lab.at/pse/}}.
We encourage the reader to use the provided dataset of embeddings in conjunction with the Projection Space Explorer for further exploration using other features.

\autoref{fig:pse_line} shows a single trajectory in the learned representation space. The trajectory starts at the right and successively moves through the middle-game in the center and ends in an endgame on the left.
It shows an oscillating behavior that is caused by the turn-taking nature of chess. Closer inspection of several trajectories reveals that different areas in the embedding space correspond to different game phases. To investigate this more closely, we use $k$-means to find 7 clusters in the full embedding space, which are colored accordingly (\autoref{fig:pse_cluster}). The clear separation of the clusters confirms that the projection into 2 dimensions via PCA maintains the clustering%
. Additionally, an analysis of the clusters shows a clear separation into different game phases: openings (\textcolor{teal}{teal}), middlegames before castling (\textcolor{orange}{orange}), white castling (\textcolor{magenta}{pink}), black castling (\textcolor{brown}{brown}), middlegames after castling (\textcolor{violet}{purple}), transitions between middle- and endgames (\textcolor{ForestGreen}{green}) and endgames with just a few pieces (\textcolor{Goldenrod}{ochre}). 

This gives a clear semantic interpretation of the game trajectory in \autoref{fig:pse_line}: 
after the opening and the first few middlegame moves, white castles, which can be inferred from the jump into the pink cluster. Afterwards, the second major jump indicates castling from black and later a transition to the endgame. The trajectory ends in the ochre region as the board only consists of pawns, the kings and a white knight, at which point black resigned.

\begin{figure}[ht!]
\centering
\includegraphics[width=0.8\linewidth]{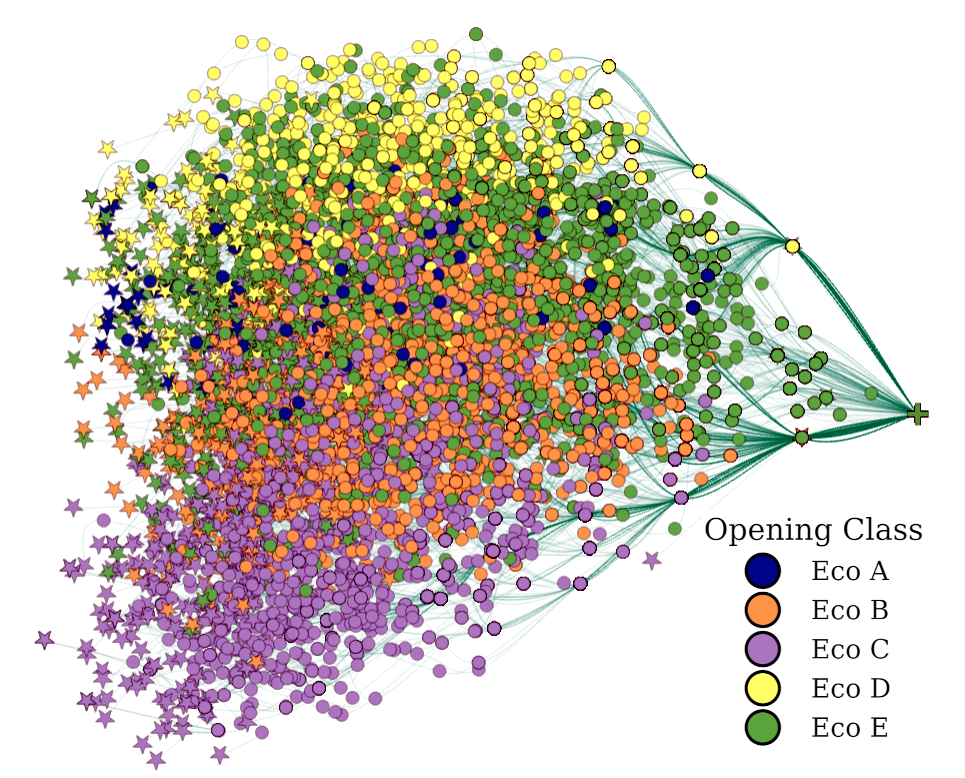}
\caption{2D projection of learned representations of chess openings, where every second state is mirrored to filter out turn oscillations. States are colored by their Eco code. We find similar openings are clustered in the embedding space.}
\label{fig:pse_mirror}
\end{figure}

To further illustrate the semantics of the embedded game trajectories, we now focus on the opening moves.
\autoref{fig:pse_mirror} shows the opening moves in a color-neutral representation to filter out turn oscillations. In other words, every position is represented from the point of view of the player to move (as opposed to from the point-of-view of white).

It can be observed that the openings are separated by color and that similar openings are clustered together. All %
queen-pawn openings are at the top half (yellow and green), whereas all king-pawn openings are at the bottom (violet and orange).

\subsubsection{ECO chess openings dataset}\label{section:Huggingface chess openings dataset}

To further explore the clustering of openings in the embedding space, we use the set of positions commonly used for the ECO chess opening classification.
The dataset consists of 3627 distinct opening board positions that can be used to classify chess games into 500 different codes from A00 to E99, where the letters A--E represent large semantic groups, such as king-pawn openings (B and C) or queen-pawn openings (D and E).

\begin{figure}[h!]
\centering
\includegraphics[width=\linewidth]{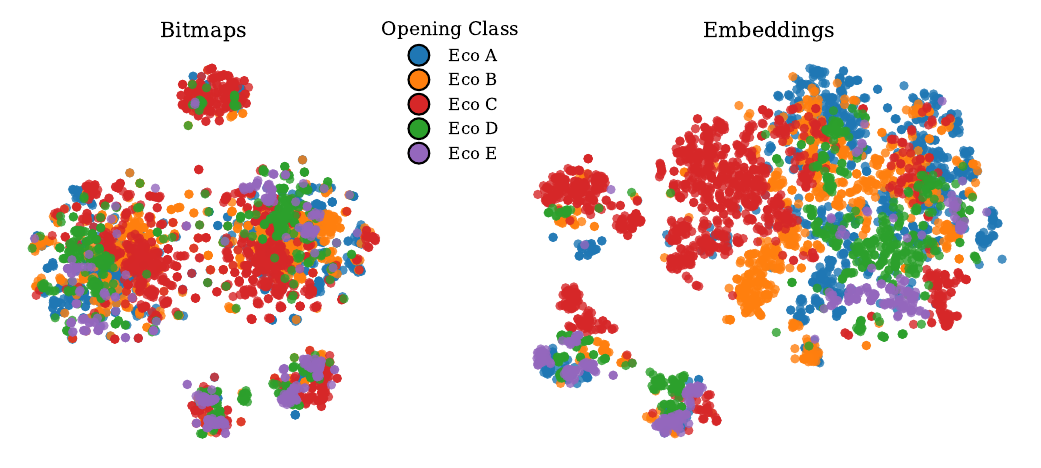}
\caption{t-SNE projected ECO chess opening positions colored by their main ECO codes \textcolor{blue}{A}, \textcolor{orange}{B}, \textcolor{red}{C}, \textcolor{ForestGreen}{D}, and \textcolor{violet}{E}. We see the original bitmaps (left) and the learned embeddings (right). The learned embeddings show a clearer separation by Eco code, highlighting the semantic structure of the space.}
\label{fig:eco_dataset}
\end{figure}

\autoref{fig:eco_dataset} shows the difference between t-SNE projections of the original bitmaps and the learned embeddings, colored by their ECO code. We use t-SNE, as PCA is not suitable for visualizing one-hot encoded representations. The left side shows the unprocessed bitmap representation, whereas the right side contains the encoded representation. The right side shows some separation of the different openings in accordance to their Eco character, while the bitmaps are less separated. We can also see that \textcolor{orange}{B} and \textcolor{red}{C} (king-pawn openings) form a clear cluster in the center (with some transpositions into other opening groups).

\subsubsection{Kaggle chess puzzle dataset}\label{section:Lichess chess puzzle dataset}

The Kaggle chess puzzle dataset 
consists of a set of chess positions annotated with several semantic keywords. They also contain the solutions, i.e., the correct line of moves, which are ignored in our experiments.

\begin{figure}[ht]
\centering
\includegraphics[width=\linewidth]{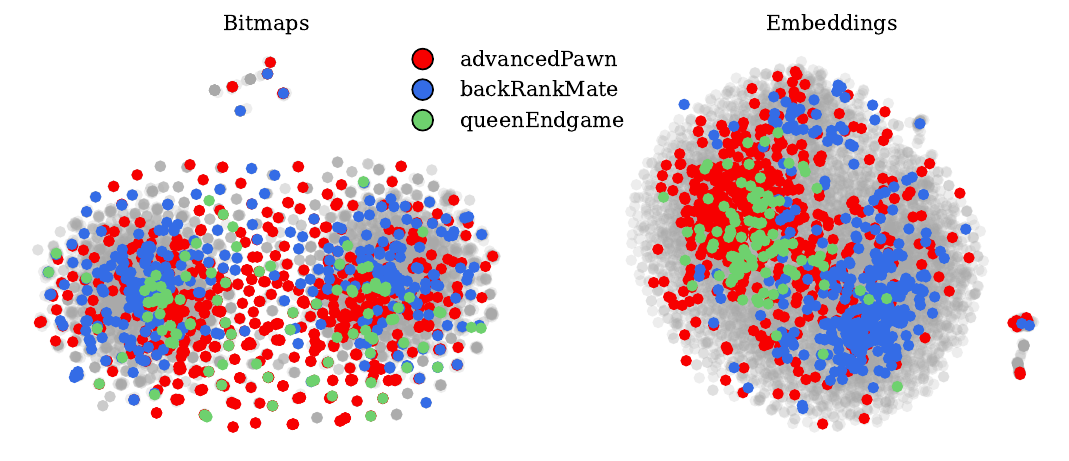}
\caption{t-SNE projections of the Lichess puzzle dataset as unprocessed bitmaps (left) and in our embedding space (right). Positions are colored according to their motifs: \textcolor{red}{advanced pawn}, \textcolor{blue}{back-rank mate}, and \textcolor{ForestGreen}{queen endgame}.}
\label{fig:puzzle}
\end{figure}

In \autoref{fig:puzzle}, we compare the original $19\times 8\times 8$ bitmap representation (left) with the embeddings learned by our model (right), both projected to 2D using t-SNE. 
The red, blue and green dots represent puzzles whose solutions involve advanced pawns, back-rank mates, and queen endgames, respectively. Positions not containing any of these motifs are colored gray.

We again find a much clearer separation in the learned embeddings than in the unprocessed bitmaps. This again confirms the semantic richness of our representation space.

\section{Conclusions}\label{section:Conclusion}

We proposed the novel self-supervised architecture \repair{}, which learns to reconstruct states in sequential problems. By masking and restoring parts of a latent sequence, the model learns to extract semantic information from sequences, providing meaningful embeddings and reconstructions for missing elements. 

We validated our architecture in chess and showed that chess positions can be mapped into a semantically rich space that enables latent state reconstruction. By decoding masked states, 
\repair{} can learn to reconstruct gaps and to provide uncertainty estimations for pieces by reasoning about their movement. This indicated an intuition of the rules and strategies of chess.

Contrary to previous approaches such as AlphaZero, where embedding spaces emerge merely as a by-product of a trained expert chess playing policy, our work focused on self-supervised training of meaningful embeddings for chess positions. 
The versatility and semantic richness of the found embeddings could be demonstrated by 
projecting a variety of chess datasets into the representation space, where we could observe meaningful clusters for different game phases, chess openings, and puzzle motifs.
Furthermore, we showed that projecting entire chess games into the representation space allows for an interactive and intuitive dissection of any game by its trajectory.

\section{Future work}\label{section:Future_Work}

Our \repair{} architecture is general and can thus be applied to any domain where data undergoes successive transformations. It would therefore be interesting to extend it to a wide range of domains that go beyond the realm of game states. While similar ideas have been explored in vision tasks, many other domains remain largely unexplored,
where the projection of sequential transformation could lead to
explainable processes.
For example, \repair{} could be used to study protein evolution. By masking and reconstructing sequences of proteins, they could potentially be mapped into a semantically rich prediction space, where different evolutions steps manifest themselves as clusters.

From an architectural perspective, it could be further investigated how incorporating task-specific prediction heads would change the resulting embeddings. For example, applying a simple feed-forward network on the latent representation that is tasked with predicting board evaluations could push the representations into different regions.

Lastly, the use of the Projection Space Explorer showed the semantic richness of the representation space. However, it is limited to two dimensions. Interacting with the paths in three-dimensional prediction space would be insightful and may reveal additional semantic clusters that are currently undiscovered.

%% file: references.bib
@article{AlphaZero,
  title={A general reinforcement learning algorithm that masters chess, shogi, and {Go} through self-play},
  author={Silver, David and Hubert, Thomas and Schrittwieser, Julian and Antonoglou, Ioannis and Lai, Matthew and Guez, Arthur and Lanctot, Marc and Sifre, Laurent and Kumaran, Dharshan and Graepel, Thore and Lillicrap, Timothy and Simonyan, Karen and Hassabis, Demis},
  journal={Science},
  volume={362},
  number={6419},
  pages={1140--1144},
  year={2018},
  doi={10.1126/science.aar6404}
}

@inproceedings{I-JEPA,
  author       = {Mahmoud Assran and
                  Quentin Duval and
                  Ishan Misra and
                  Piotr Bojanowski and
                  Pascal Vincent and
                  Michael G. Rabbat and
                  Yann LeCun and
                  Nicolas Ballas},
  title        = {Self-Supervised Learning from Images with a Joint-Embedding Predictive
                  Architecture},
  booktitle    = {Proceedings of the {IEEE/CVF} Conference on Computer Vision and Pattern Recognition (CVPR)},
  address = {Vancouver, BC, Canada},
  pages        = {15619--15629},
  publisher    = {{IEEE}},
  year         = {2023},
  doi          = {10.1109/CVPR52729.2023.01499},
}

@article{lecun2022path,
  title={A path towards autonomous machine intelligence},
  author={LeCun, Yann},
  journal={Open Review},
  volume={62},
  number={1},
  pages={1--62},
  year={2022},
  note={version 0.9.2}
}

@inproceedings{BERT,
  author       = {Jacob Devlin and
                  Ming{-}Wei Chang and
                  Kenton Lee and
                  Kristina Toutanova},
  editor       = {Jill Burstein and
                  Christy Doran and
                  Thamar Solorio},
  title        = {{BERT:} {P}re-training of Deep Bidirectional Transformers for Language
                  Understanding},
  booktitle    = {Proceedings of the 2019 Conference of the North American Chapter of
                  the Association for Computational Linguistics: Human Language Technologies ({NAACL-HLT}), Volume 1},
  address = {Minneapolis, MN, USA)},
  pages        = {4171--4186},
  publisher    = {ACL},
  year         = {2019},
  doi          = {10.18653/V1/N19-1423}
}

@inproceedings{MAE,
  author       = {Kaiming He and
                  Xinlei Chen and
                  Saining Xie and
                  Yanghao Li and
                  Piotr Doll{\'{a}}r and
                  Ross B. Girshick},
  title        = {Masked Autoencoders Are Scalable Vision Learners},
  booktitle    = {Proceedings of the {IEEE/CVF} Conference on Computer Vision and Pattern Recognition ({CVPR})},
  address = {New Orleans, LA, USA},
  pages        = {15979--15988},
  publisher    = {{IEEE}},
  year         = {2022},
  doi          = {10.1109/CVPR52688.2022.01553},
}

@article{Chess2Vec,
  author       = {Berk Kapicioglu and
                  Ramiz Iqbal and
                  Tarik Koc and
                  Louis Andre and
                  Katharina Volz},
  title        = {Chess2vec: Learning Vector Representations for Chess},
  journal      = {arXiv},
  volume       = {abs/2011.01014},
  year         = {2020},
  url          = {https://arxiv.org/abs/2011.01014},
}

@article{SqueezeAndExcitation,
  author       = {Jie Hu and
                  Li Shen and
                  Samuel Albanie and
                  Gang Sun and
                  Enhua Wu},
  title        = {Squeeze-and-Excitation Networks},
  journal      = {{IEEE} Transactions on Pattern Analysis and Machine Intelligence},
  volume       = {42},
  number       = {8},
  pages        = {2011--2023},
  year         = {2020},
  doi          = {10.1109/TPAMI.2019.2913372}
}

@article{RoPE,
  author       = {Jianlin Su and
                  Murtadha H. M. Ahmed and
                  Yu Lu and
                  Shengfeng Pan and
                  Wen Bo and
                  Yunfeng Liu},
  title        = {{RoFormer}: {E}nhanced transformer with Rotary Position Embedding},
  journal      = {Neurocomputing},
  volume       = {568},
  pages        = {127063},
  year         = {2024},
  doi          = {10.1016/J.NEUCOM.2023.127063},
}

@inproceedings{BYOL,
author = {Grill, Jean-Bastien and Strub, Florian and Altch\'{e}, Florent and Tallec, Corentin and Richemond, Pierre H. and Buchatskaya, Elena and Doersch, Carl and Pires, Bernardo Avila and Guo, Zhaohan Daniel and Azar, Mohammad Gheshlaghi and Piot, Bilal and Kavukcuoglu, Koray and Munos, R\'{e}mi and Valko, Michal},
title = {Bootstrap your own latent: A new approach to self-supervised Learning},
year = {2020},
isbn = {9781713829546},
publisher = {Curran Associates Inc.},
address = {Red Hook, NY, USA},
booktitle = {Proceedings of the 34th International Conference on Neural Information Processing Systems (NeurIPS)},
articleno = {1786},
numpages = {14},
location = {Vancouver, BC, Canada},
_series = {NeurIPS '20}
}

@inproceedings{DINO,
  title={Emerging Properties in Self-Supervised Vision Transformers},
  author={Mathilde Caron and Hugo Touvron and Ishan Misra and Hervé Jégou and Julien Mairal and Piotr Bojanowski and Armand Joulin},
  booktitle = {Proceedings of the IEEE/CVF International Conference on Computer Vision (ICCV)},
  year={2021},
  pages={9630--9640},
}

@misc{Stockfish,
  title       = {Stockfish Chess Engine},
  author      = {Tord Romstad and Marco Costalba and Joona Kiiski and Gary Linscott},
  year        = {2023},
  version     = {15},
  url         = {https://stockfishchess.org/}
}

@misc{lc0,
  author = {{Leela Chess Zero team}},
  title = {Leela Chess Zero},
  url = {http://lczero.org/},
  urldate = {2026-01-13},
  year = {2026}
}

@article{Takase2021,
  title={Lessons on Parameter Sharing across Layers in Transformers},
  author={Sho Takase and Shun Kiyono},
  journal={arXiv},
  year={2021},
  volume={abs/2104.06022},
  url = {https://arxiv.org/abs/2104.06022},
}

@inproceedings{Dabre2019,
author = {Dabre, Raj and Fujita, Atsushi},
title = {Recurrent stacking of layers for compact neural machine translation models},
year = {2019},
isbn = {978-1-57735-809-1},
publisher = {AAAI Press},
_url = {https://doi.org/10.1609/aaai.v33i01.33016292},
doi = {10.1609/aaai.v33i01.33016292},
booktitle = {Proceedings of the 33rd AAAI Conference on Artificial Intelligence},
articleno = {772},
numpages = {8},
location = {Honolulu, Hawaii, USA},
_series = {AAAI'19/IAAI'19/EAAI'19}
}

@article{VCReg,
author = {Zhu, Jiachen and Shwartz-Ziv, Ravid and Chen, Yubei and Lecun, Yann},
journal={arXiv},
year = {2023},
month = {06},
pages = {},
title = {Variance-Covariance Regularization Improves Representation Learning},
doi = {10.48550/arXiv.2306.13292},
url = {https://arxiv.org/abs/2306.13292},
volume={abs/2306.13292},
}

@article{PCA,
author = {Karl Pearson},
title = {{LIII}. {O}n lines and planes of closest fit to systems of points in space},
journal = {The London, Edinburgh, and Dublin Philosophical Magazine and Journal of Science},
volume = {2},
number = {11},
pages = {559--572},
year = {1901},
publisher = {Taylor \& Francis},
doi = {10.1080/14786440109462720},
}

@article{t-SNE,
  title={Visualizing Data using t-{SNE}},
  author={Laurens van der Maaten and Geoffrey E. Hinton},
  journal={Journal of Machine Learning Research},
  year={2008},
  volume={9},
  pages={2579--2605},
}

@inproceedings{Maia-2,
author = {Tang, Zhenwei and Jiao, Difan and McIlroy-Young, Reid and Kleinberg, Jon and Sen, Siddhartha and Anderson, Ashton},
title = {Maia-2: a unified model for human-AI alignment in chess},
year = {2024},
isbn = {9798331314385},
publisher = {Curran Associates Inc.},
address = {Red Hook, NY, USA},
booktitle = {Proceedings of the 38th International Conference on Neural Information Processing Systems (NeurIPS)},
articleno = {659},
numpages = {26},
location = {Vancouver, BC, Canada},
_series = {NIPS '24}
}

@article{PSE,
    title = {Projection Path Explorer: Exploring Visual Patterns in Projected Decision-Making Paths},
    author = {Andreas Hinterreiter and Christian A. Steinparz and Moritz Heckmann and Holger Stitz and Marc Streit},
    journal = {ACM Transactions on Interactive Intelligent Systems},
    doi = {10.1145/3387165},
    _url = {https://dl.acm.org/doi/10.1145/3387165},
    volume = {11},
    number = {3–4},
    pages = {Article 22},
    year = {2021}
}

@misc{2025_constrastive_chess,
      title={Learning to Plan via Supervised Contrastive Learning and Strategic Interpolation: A Chess Case Study}, 
      author={Andrew Hamara and Greg Hamerly and Pablo Rivas and Andrew C. Freeman},
      year={2025},
      eprint={2506.04892},
      archivePrefix={arXiv},
      primaryClass={cs.CV},
      url={https://arxiv.org/abs/2506.04892}, 
}
